%% file: main.tex
\documentclass[letterpaper, 10 pt, conference]{sty/ieeeconf}

\IEEEoverridecommandlockouts
\usepackage{amsmath,amssymb,amsfonts,siunitx}
\usepackage{graphicx,epsfig,layouts}

\let\labelindent\relax
\usepackage{enumitem}
\usepackage{comment}
\usepackage{tabularx}
\usepackage{booktabs}
\usepackage{colortbl}

\usepackage[dvipsnames]{xcolor}
\usepackage[font=small,labelfont=bf]{caption}
\usepackage[position=b]{subcaption}
\usepackage[utf8]{inputenc}

\usepackage{flushend}   
\usepackage{url}
\usepackage{cleveref}
\usepackage{tikz}
\usetikzlibrary{calc,math}
\usepackage{sty/custom}

\title{LF: Online Multi-Robot Path Planning Meets Optimal Trajectory Control}
\author{Ajay Shankar$^*$, Keisuke Okumura$^*$, Amanda Prorok
\thanks{
The authors are with the University of Cambridge, UK.
KO is also with National Institute of Advanced Industrial Science and Technology (AIST), Japan.
$^*$~Equal contribution; ordering decided by three tosses of a fair £1 coin.
Emails: {\tt\{as3233,ko393,asp45\}@cst.cam.ac.uk}.
}
\thanks{
This work was supported in part by European Research Council (ERC) Project 949940 (gAIa), and by JSPS Overseas Research Fellowship and JST ACT-X Grant Number JPMJAX22A1.
}
}

\begin{document}
\maketitle
\input{abs}
\input{content}
\bibliographystyle{sty/IEEEtran}
\bibliography{sty/ref-macro,ref}
\end{document}

%% file: abs.tex
\begin{abstract}
We propose a multi-robot control paradigm to solve point-to-point navigation tasks for a team of holonomic robots with access to the full environment information.
The framework invokes two processes asynchronously at high frequency:
\emph{(i)}~a centralized, discrete, and full-horizon planner for computing collision- and deadlock-free paths rapidly, leveraging recent advances in multi-agent pathfinding (MAPF), and
\emph{(ii)}~dynamics-aware, robot-wise optimal trajectory controllers that ensure all robots independently follow their assigned paths reliably.
This hierarchical shift in planning representation from \emph{(i)}~discrete and coupled to \emph{(ii)}~continuous and decoupled domains enables the framework to maintain long-term scalable motion synthesis.
As an instantiation of this idea, we present \LF, which combines a fast state-of-the-art MAPF solver (LaCAM), and a robust feedback control stack (Freyja) for executing agile robot maneuvers.
\LF provides a robust and versatile mechanism for lifelong multi-robot navigation even under asynchronous and partial goal updates, and adapts to dynamic workspaces simply by quick replanning.
We present various multirotor and ground robot demonstrations, including the deployment of 15 real multirotors with random, consecutive target updates while a person walks through the operational workspace.
\end{abstract}

\medskip
\noindent
\textbf{Project page: \url{https://proroklab.github.io/lf}}

%% file: content.tex
\section{Introduction}
Efficiently navigating a team of robots to their respective destinations---whilst maintaining responsiveness, and avoiding collisions, deadlocks, or livelocks---is a crucial skill for any multi-robot team.
Deploying such a team reliably in practical situations often faces numerous challenges in the complexity of planning and control, assumptions on dynamics, non-stationary environments, uncertainties from inter-robot or human-robot interactions, and imperfections in deployment and synchronization~\cite{gielis2022critical}.
These factors make a single-shot, open-loop planning strategy impractical.
Ideally, a fast, reactive motion planner is implemented to maintain team-level coordination, tightly coupled with a low-level trajectory controller for precise individual motion control.

Coupled planning and control in continuous spaces, performed over the team's joint configuration space, typically incurs a high computational cost.
Instead, decoupled counterparts that rely on local observations and communication have received considerably more attention in recent years~\cite{wang2017safety,zhou2017fast,luis2020online,tordesillas2021mader,zhou2022swarm}.
While their computational efficiency makes them excel at coordinated behaviors in the short-term, decoupled methods generally lack long-term coordination guarantees, such as preventing local deadlocks, especially in dense and constrained environments.

\input{figs/top}

In parallel, recent advances in coordinated path planning in discrete domains, framed as multi-agent pathfinding (MAPF) algorithms, have demonstrated remarkable scalability, handling systems with thousands of agents in short planning timeframes~\cite{li2022mapf,okumura2023lacam2}.
While these algorithms typically employ abstractions such as discrete synchronous actions within gridworlds, the computational efficiency of such scalable MAPF algorithms open the door to a new paradigm where planning is no longer a one-shot open-loop task.
Combined with recent advancements in the maturity of efficient on-robot trajectory control stacks, we can now naturally \textit{embed} MAPF routines online within a control loop.

Two key developments thus motivate this step-change:
\emph{(i)}~the emergence of ultra-fast MAPF algorithms that enable online replanning, and
\emph{(ii)}~the availability of efficient low-level optimal controllers capable of precise trajectory execution for individual robots.
We propose to run team-level, full-horizon MAPF algorithms online (e.g., \SIrange[range-phrase = --,range-units = single]{5}{10}{Hz}), and use the latest system state to continuously update plans in real-time.
Concurrently, decoupled, robot-wise, higher-frequency (\SIrange{50}{100}{Hz}) trajectory controllers account for real-world dynamics and track the high-level plans with high fidelity.
As an instantiation, we develop the \lf framework for holonomic multi-robot control, which combines: \emph{(i)}~a state-of-the-art coupled MAPF algorithm called LaCAM~\cite{okumura2024lacam3}, and \emph{(ii)}~a decoupled, low-level optimal trajectory control system called Freyja~\cite{shankar2021freyja}.

This hierarchical scheme is particularly attractive as it addresses the limitations of fully decoupled approaches (i.e., lack of long-term coordination) without sacrificing real-time responsiveness.
The rationale is that, at each replanning, \LF derives full-horizon paths that ensure the robots can reach their destination without collisions, deadlocks, or livelocks---as long as the robots can track their paths, they are guaranteed to coordinate.
For severe deviations and dynamic scenarios, \LF's fast replanning ensures feasibility.
Indeed, \cref{fig:top} demonstrates 10-agent navigation with dynamic obstacles, with \SI{5}{\hertz} MAPF run on a consumer-level laptop.
Simulations also confirm the planner's ability to continuously replan at \SI{10}{\hertz} for dense teams of up to 32 robots.

As indicated in \cref{table:characterization}, this work bridges a key technological gap that currently limits a close-knit integration of team-level path planners and robot controllers.
We show that with reasonable adaptations to both, this is now attainable, and can be extended to support a wide range of multi-robot use-cases (see supplementary video).
These include synchronous, asynchronous, and sporadic replanning, lifelong navigation tasks~\cite{li2021lifelong,zhou2022swarm} and multirotor swarm deployments with a pedestrian.
The complete implementation developed in C++ and ROS2, and videos are available on the website.

\section{Target Systems}
\label{sec:target_sys}
Our objective is to solve point-to-point navigation tasks for holonomic robot fleets in $\mathbb{R}^2$ or $\mathbb{R}^3$ with access to full state and map.
Specifically, we target robotic systems that exhibit differentially flat dynamics, which enables planning in the (potentially joint) state space rather than in action space.
Examples include ground-robot platforms such as the Cambridge Robomaster~\cite{blumenkamp2024cambridge}, or aerial platforms such as the Crazyflie~\cite{giernacki2017crazyflie}, Sanity~\cite{woo2025sanity} etc, the main application focus of this paper.
The task is to continuously navigate multiple robots, each with its own destination set by an external task assignee, possibly updated asynchronously at any point during the operation.
We use `agent' and `robot' interchangeably.

\section{Related Work}
The literature on decoupled, potentially decentralized schemes to solve point-to-point navigation in continuous spaces is rich, especially for aerial swarms~(e.g.,~\cite{wang2017safety,zhou2017fast,luis2020online,tordesillas2021mader}).
Our interest lies in synthesizing guaranteed coordination, challenging to establish with such decoupled methods.
There are coupled, numerical optimization-based control schemes, but these are either open-loop controls with offline pre-computed trajectories~\cite{augugliaro2012generation,honig2018trajectory,park2020efficient,moldagalieva2024db,pan2024hierarchical}, or closed-loop controls but with a limited planning horizon~\cite{soria2021predictive,adajania2023amswarm,adajania2024amswarmx,tajbakhsh2024conflict}.
The former approaches cannot respond to dynamic environments with moving obstacles or on-demand target assignments, while the latter sacrifices the synthesis of long-term optimal coordination which is the most appealing property of coupled methods.
As summarized in \cref{table:characterization},%
\footnote{
Tight coordination refers to the ability to advance the mission in densely populated situations, while long coordination refers to the ability to avoid congestion or dead ends as a team a priori.
\LF is inferior in smoothness and aggressiveness because the high-level planner lacks awareness of the finer-grained low-level dynamics, as discussed later.
}
we develop a new category employing real-time, coupled, and full-horizon planning, in the sense that we generate a trajectory from start to destination, with the help of computationally lightweight MAPF techniques that have emerged in recent years.

MAPF has been extensively studied since the 2010s, primarily targeting gridworld, discrete-time representations~\cite{ma2022graph}.
Recent scalable methods can suboptimally solve thousands of agents in seconds~\cite{li2022mapf,okumura2023lacam}.
Besides, there are anytime refinement schemes~\cite{li2021anytime,okumura2021iterative} which rapidly refine solution quality over time.
These advances enable the realization of \LF, which \textit{embeds} the MAPF within a feedback loop at a higher frequency.

This feedback planning differs from conventional MAPF execution styles, which solve MAPF continuously but at lower frequency (e.g. every \SIrange[range-phrase = --,range-units = single]{1}{10}{s}), such as online replanning at each discrete timestep~\cite{vsvancara2019online,li2021lifelong}, post-processing MAPF plans with liveness guarantees~\cite{honig2016multi,mannucci2021provably,berndt2023receding}, or offline planning robust to uncertainty~\cite{atzmon2020robust,okumura2023offline}.
Some studies include online MAPF replanning during robot deployment~\cite{hou2022enhanced,park2023decentralized,pan2024hierarchical}, but they limit the planning horizon or subdivide the planning problem into small teams to reduce the computational burden, thus forgoing the advantages of coupled planning.
Our approach is much more intuitive, without such tweaks;
we aim to achieve seamless and robust multi-robot control by running MAPF more frequently.

\input{table/characterization}
\input{figs/arch}

\section{Architecture}

Given the current system state and the target positions assigned to each robot, \lf periodically and asynchronously calls the following two sub-processes:
\emph{(i)}~a high-level coordination planner that efficiently solves MAPF, and
\emph{(ii)}~a low-level optimal control mechanism that continuously processes the robots' trajectories based on the latest instructions, a sequence of waypoints, provided by the high-level planner.
The integration of the two, which operate in very different representations, is readily justified for holonomic robots that exhibit differentially flat dynamics and enable planning directly in Euclidean space. 
Typically, the high-level planner operates at high frequency (e.g., $5$--$10$ Hz) offboard, enabling it to adapt promptly to dynamic, noisy, and imperfect environments.
The low-level controller runs at much higher frequencies (e.g., $50$--$100$ Hz), potentially onboard, depending on the requirements of the robotic system, ensuring precise and smooth control of individual robots.
\lf adapts LaCAM and Freyja as the embodiment of this concept.
\Cref{fig:arch} illustrates the architecture.

The concept is elegant and compositional in that the two components are naturally required for any multi-robot team, and yet, there are specific considerations we make to each.
The following two sections highlight these aspects.

\section{LaCAM in Continuous Spaces}
\subsection{Preliminaries and Considerations}
MAPF is a combinatorial search problem for assigning collision-free paths to multiple agents on a geometric graph, assuming that all agents take actions synchronously.
This classical MAPF formulation is reduced to single-agent pathfinding on a graph consisting of \emph{configurations}, the joint state of all agents, each marking a location for each agent.
Two configurations are neighboring if any agent can move from one to the other within one timestep, and there are no inter-agent collisions during this transition.
This connectivity defines a graph structure; thus, any search algorithm, such as depth-first search (DFS), can derive a solution to MAPF.

In practice, however, such exhaustive search methods only work when the number of agents is small because of the exponential growth of the number of successor configurations.
LaCAM primarily uses DFS, but mitigates this impracticality by generating a successor sequentially, while imposing constraints on each generation to guarantee that all successors will eventually appear.
This lazy successor generation dramatically reduces the computational burden, resulting in a quick and scalable approach to MAPF.

A key challenge in deploying MAPF algorithms directly over continuous spaces is the traditional grid-world representation they employ for discrete search.
To generate feasible paths, the search within LaCAM in \LF uses \emph{rotational} motion primitives on an agent-wise coarse roadmap, which is partially inspired from~\cite{cohen2019optimal}.
The following describes the key considerations in making this adjustment.

\subsection{Problem Definition}
We define the MAPF problem in continuous spaces as follows.
The problem instance consists of a closed workspace $\W \subset \mathbb{R}^3$, obstacle regions $\W\sub{obs} \subset \W$, a team of holonomic and homogeneous robots $A = \{1, 2, \ldots, n\}$, their current position $\Q: A \mapsto \W$ and target $\T: A \mapsto \W$, agent radius $r\sub{agent} \in \mathbb{R}_{>0}$, and target radius $r\sub{target} \in \mathbb{R}_{>0}$.
A planner aims to derive collision-free paths, assuming single integrator dynamics with constant speed motion, with a maximum travel distance $d\sub{travel} \in \mathbb{R}_{>0}$ per unit time.
Specifically, a solution assigns each agent $i \in A$ a sequence of geometric locations $[p_i^0, p_i^1 \ldots, p_i^T]$, $p_i^k \in \W$, that satisfies:
{\setlength{\jot}{0.25ex}
\begin{align*}
    p_i^0 &= \Q(i) \\
    \| p_i^T - \T(i) \| &\leq r\sub{target} \\
    \| p_i^{k} - p_i^{k+1} \| &\leq d\sub{travel} \\
    (1-\alpha) p_i^k + \alpha p_i^{k+1} \equiv p_i^{k \rightarrow k+1} &\not\in \W\sub{obs} \\
    \| p_i^{k \rightarrow k+1} - p_j^{k \rightarrow k+1} \| &\ge 2r\sub{agent}
\end{align*}
}%
where $0 \leq \alpha \leq 1$.
The last two prevent collisions.

\subsection{Construction of Agent-Wise and Coarse Roadmap}
The first step in solving the above problem is to construct a roadmap $G_i=(V_i, E_i)$ in \W for each agent $i \in A$.
To represent $\W\sub{obs}$, \lf uses OctoMap~\cite{hornung2013octomap}, a popular 3D environment representation in robotics.
Then $G_i$ is constructed for each replanning, with collision checking performed using the Flexible Collision Library (FCL)~\cite{pan2012fcl}.
Each vertex is assigned a cost-to-go value to the goal location $\T(i)$, denoted as $\phi_i: V_i \mapsto \mathbb{R}_{\geq 0}$, calculated using the Dijkstra algorithm.

Unlike the common use of the roadmap, which performs a direct discrete search on $G_i$, our purpose is to approximate the gradient to the destination given the query position $p \in \mathbb{R}^3$.
Let denote the neighboring location sets on the roadmap $\N(p, V_i)$, which is efficiently identified by the radius search on the k-d tree on $V_i$.
\LF then calculates the obstacle-aware gradient to $\T(i)$, denoted as $\nabla_i(p)$:
\begin{align}
\nabla_i(p) = \funcname{norm}\left( \sum_{v \in \N(p, V_i)} \left(\phi_i(v) - \bar{\phi} \right)\cdot(v - p) \right)
\end{align}
where \funcname{norm} is a normalizing operation, and,
\begin{align}
\bar{\phi} = \frac{1}{|\N(p, V_i)|} \sum_{v \in \N(p, V_i)} \phi_i(v).
\end{align}
Without considering inter-robot collisions, each robot $i$ will eventually reach its destination if it follows this gradient.

\subsection{Search with Rotational Motion Primitives}
Let $R(\nabla_i(p))$ be a rotation matrix that transforms a unit vector $[1, 0, 0]^\top$ being aligned with $\nabla_i(p)$.
To perform discrete path planning in continuous spaces, we define an implicit search graph for agent $i$ with
\begin{equation}
\S_i(p) = \{p + d\sub{travel} \cdot R(\nabla_i(p)) e \mid e \in \M\}
\end{equation}
where $\M$ is a set of (unit) motion primitives
$\{[\pm 1, 0, 0]^\top$, $[0, \pm 1, 0]^\top$, $[0, 0, \pm 1]^\top$, $[0, 0, 0]^\top \}$.

With the notion of $\S_i$, it is possible to define a finer-grained graph structure $\G_i$ that spans over $\W \setminus \W\sub{obs}$, starting from the current location $\Q(i)$.
We adapt a LaCAM implementation, originally developed for four-connected grid MAPF, to perform an exhaustive search over the joint state space for all agents, i.e., $\G_1 \times \G_2, \ldots \times \G_n$, to derive a solution from the current configuration $\Q$.
The implicit search graph $\G_i$ is not constructed a priori, rather, it is constructed gradually as the search progresses.
Static obstacle collision checking uses OctoMap and FCL, while inter-agent one uses a simple geometric calculation based on spherical shapes in the problem definition.
Unlike the discrete search, in the continuous state representation, there is no exact state duplication whose detection is critical to continue the search.
Thus we consider a search node on the configuration for $Q' \in \mathbb{R}^{3n}$ to be duplicated if there is a configuration geometrically close to $Q'$ based on their agent-wise distances.

Note that the search graph $\G_i$ is different from the coarser roadmap $G_i$.
This hierarchical workspace representation is designed to derive smoother solutions with less computation to meet real-time usage.
It is critical to keep the size of motion primitives \M to a minimum to avoid a large branching factor in the path planning.
However, a vanilla construction $\M$ with unit vectors can only derive zigzag paths, making it difficult for the controller to track the paths smoothly.
This motivates the introduction of rotation for the generation of successor states.
The gradient information can be retrieved with $G_i$, which must also be minimal because of its construction cost with the overhead of collision checking.
The scheme employed meets these considerations.

\subsection{Anytime Refinement}
LaCAM continues to refine solutions as time allows after the initial solution discovery.
Unlike LaCAM$^\ast$~\cite{okumura2023lacam2}, an asymptotically optimal variant with search-tree rewiring, \lf employs branch-and-bound refinement.
It also incorporates advanced techniques from~\cite{okumura2024lacam3}, including Monte-Carlo successor generation and dynamic solution updates obtained from large neighborhood search~\cite{li2021anytime,okumura2021iterative}.
In addition, path smoothing techniques~\cite{geraerts2007creating} are introduced;
from a solution path $\pi_i = [p^i_0, p^i_1, \ldots, p^i_T]$, our smoothing implementation first generates another sequence $\pi_i\suf{skip}$, skipping some intermediate locations in $\pi_i$ with a small probability.
Then a new candidate $\pi_i\suf{smooth}$ is sampled from the linear interpolation fit of $\pi_i\suf{skip}$.
If $\pi_i\suf{smooth}$ has no collisions with paths of other agents, $\pi_i$ is replaced by $\pi_i\suf{smooth}$.

\input{figs/navigation_performance}

\subsection{Feedback Loop Integration}
With the combination of the above techniques, the high-level planner delivers high-quality coordinated paths to the low-level controller quickly.
The feedback loop is closed by the interface/controllers updating the initial configuration $\Q$ and invoking the planner when required.
However, as robot tracking is not perfectly accurate, $\Q$ may be infeasible with, e.g. including inter-agent collisions or collisions with static obstacles, which prevents the combinatorial search from deriving a solution.
In such cases, \LF adds small amounts of geometric noise to $\Q$ until it becomes feasible.

To maximize the solution quality within severe computation time limit, partially inspired from~\cite{zhang2024planning}, \LF reuses the solution from the previous planning, $\pi\suf{prev} = [\Q^0, \Q^1, \ldots, \Q^T]$, if the tracking deviation is negligible and $\pi\suf{prev}$ is still feasible.
In particular, given a new query with $\Q\suf{new}$, if there is a configuration $\Q^k \in \pi\suf{prev}$ such that $\max_{i \in A} \| \Q^k[i] - \Q\suf{new}[i]\|$ is below a threshold, $\Q\suf{new}$ is replaced by $\Q^k$ and LaCAM immediately starts anytime refinement using a seed solution $[\Q^{k}, \Q^{k+1}, \ldots, \Q^T]$.
If not, LaCAM simply starts the search from scratch.
This eliminates the costly time of initial solution discovery in most cases, resulting in better high-level plans by the planning deadline.

\section{Freyja}
Freyja is a model-based optimal non-linear feedback control stack for executing fast and agile robot maneuvers;
the following description focuses on multirotor platforms but the target systems are not restricted to these.
Our implementation is setup as a collection of three primary ROS2
`nodes' that perform state estimation, state regulation over a given trajectory, and vehicle communication handling, as well as an additional interface element as part of \LF{}.
Freyja provides several configurable options for each module that are tunable for specific instances and use-cases.
In \lf, we use a Kalman filter to estimate the 6-DoF state, along with its first and second derivatives, using pose measurements from a motion-capture system.
A Linear Quadratic Gaussian (LQG) controller is used as a state regulator to generate control actions that drive the robot along a path generated from LaCAM.
Note that due to the differentially flat dynamics of our target systems, it is possible to perform planning in $\mathbb{R}^3$.
Freyja specifically exploits this property to then map the feedforward-linearized control actions~\cite{greeff2018flatness} into the non-linear action space defined by the target attitude and thrust vector for a multirotor.
The LQG implementation is extremely robust for a wide range of flight regimes, and offers high computational efficiency, e.g., it can be run at over \SI{200}{Hz} on a Raspberry Pi Zero.
For larger problems with more constraints, Freyja also supports a versatile model predictive control architecture by interfacing with QP-solver frameworks such as OSQP~\cite{osqp}.

\subsection{Trajectory Control}
For multirotor platforms, Freyja can regulate continuous state trajectories,
$\boldsymbol{x} \in \mathbb{R}^{10}$
with
$\boldsymbol{x}\equiv[p, \dot{p}, \ddot{p}, \psi]^\top$, where $p$ represents the position in $\mathbb{R}^3$ and $\psi$ is the orientation (yaw).
Specifically, in our LQG implementation, we regulate a projected linear system that evolves according to
$\dot{\boldsymbol{x}} = A\boldsymbol{x} + B\boldsymbol{u}$ such that we can compute a closed-form control input, $\boldsymbol{u} = -K_{\mathrm{fb}}\boldsymbol{x} + K_{\mathrm{ff}}\ddot{p}$ that effects accelerations on the system.
The feedback gain matrix, $K_{\mathrm{fb}}$, is precomputed by solving the infinite-horizon discrete algebraic Ricatti equation, and $ K_{\mathrm{ff}}$ is generally a tunable parameter.
For teams comprised of heterogeneous robots, these precomputed matrices will be different and can be obtained offline (once).

Three key elements are now essential to \LF{}.
First, the control architecture of Freyja is relatively light-weight, compiled into small binaries, and can even be distributed onboard each robot independently.
We also enable several per-robot components, such as disturbance observers that depend only on local dynamics and help improve tracking accuracy.
Second, the acceleration-space control vector, $\boldsymbol{u}$, is transformed into a non-linear space that represents the specific input space of the robot.
This is done by inverting the dynamics projection that enabled the differentially flat mapping.
For instance, for multirotors, we can define the non-linear map
$f^{-1}: \mathbb{R}^4\mapsto\mathbb{R}^4$ that projects the actual plant inputs $[\phi, \theta, \dot{\psi}, Tc]^\top$ to the control $\boldsymbol{u}$, where $\phi, \theta$ and $Tc$ denote roll, pitch and collective thrust, respectively.
By obtaining these using $f(\boldsymbol{u})$, we unlock finer control over trajectory tracking and sidestep linearization effects.
Third, we describe the transformation that converts LaCAM outputs to trajectories tracked by Freyja as follows.

\subsection{Discrete Paths to Continuous Trajectories}
Since a path, $\pi$, generated by LaCAM is continuous in geometry but is represented as discrete-time samples, it lacks continuity as a trajectory, $\boldsymbol{x}$.
Recall also that $\pi$ is comprised of $p^k$ terms only, and is often much sparser than acceptable.
Thus in \LF, Freyja implements an additional intermediate step that refines $\pi \rightarrow \boldsymbol{x}$.
This is done using a linear 2nd-order interpolation
that enables linearization over points on a trajectory (i.e., a point with velocity and acceleration vectors) as opposed to individual points in space, thus producing smoother motions.
We arbitrarily set $\psi=0$, and compute interpolated $\dot{p}, \ddot{p}$ at Freyja's loop rate.
Doing so has $O(T)$ cost, and in practice, is fully decoupled for each agent.

\section{Experiments}
Our framework supports a close integration of the planner within a lower-level control loop. In our current implementation, the planner can be triggered upto \SI{20}{Hz} -- this can be synchronous (for all agents simultaneously) or asynchronous, periodic (fixed-rate) or sporadic (event-triggered, based on goal assignment, current tracking performance, or detection of new obstacles).
We deploy \LF{} with teams of indoor multi-robot testbed platforms~\cite{woo2025sanity,blumenkamp2024cambridge}, primarily off-the-shelf multirotors.
All experiments are conducted on a laptop equipped with an M3 Max Apple MacBook and \SI{64}{\giga\byte} RAM.
The localization uses a motion capture system that covers about \qtyproduct[product-units=bracket-power]{7 x 5 x 1.8}{\metre}.
In the following subsections, we will showcase the ability of our framework to handle various complex environments, obstacles, dynamism, limited time resources, and scalability to a large number of agents.
Since \LF{} is built upon a centralized coupled planner, we will also juxtapose its solutions against a fast state-of-the-art centralized, \textit{decoupled} method called AMSwarm~\cite{adajania2023amswarm}.

\subsection{Navigation Performance}
\label{sec:navigation_performance}
We begin by evaluating \LF's navigation capabilities using simple scenarios, but with real multirotors.
Our main interest is quantifying tracking errors between intended and actual trajectories---since the planner always returns feasible solutions without any collisions, deadlocks, or livelocks,
within reasonable tolerance, \LF will navigate all robots without any fatal events.
Experiments are conducted with repeated missions over a minute, where new goals are assigned to the team as soon as \textit{all} agents have reached their current goals, which we will later call the \emph{synchronous} scenario.

\input{figs/planner_anytime}
\input{figs/planner_test}
\input{figs/juxt}

\Cref{fig:navigation_performance}a shows the distribution of tracking errors for five multirotors deployed with varying MAPF planning frequencies and tracking speeds.
In the figure, `One-shot' corresponds to the nai\"ve execution style that may be adopted traditionally with slower planners that cannot be retriggered during the mission, and thus cannot deal with severe tracking failures or dynamic obstacles.
This establishes a baseline tracking capability, typically \SIrange{0.1}{0.15}{\meter}, which is nearly half the robot size parameter.
When using \LF{} with online replanning, either at \SI{5}{\hertz} or \SI{10}{\hertz}, we observe a slight improvement in tracking performance, despite frequent updates to the high-level plan.
This is owing to the robustness of the robot-wise trajectory controller and the consistency of MAPF plans over replanning trials.

\Cref{fig:navigation_performance}(b, c) visualizes results from a ten robot deployment over \SI{2}{\minute} with four obstacles that change their location slowly over time.
\LF{} replaning is triggered at \SI{5}{\hertz}.
The target for each agent is randomly sampled from a 2D circle with a radius of \SI{2}{\meter}.
\Cref{fig:top} also presents a snapshot of one mission.
The deviation tends to be greater than \cref{fig:navigation_performance}a, because dynamic obstacles sometimes force \LF to override the current plan entirely,
and then it takes time to stabilize to a smooth solution (c.f. \cref{sec:stress-test}).

\subsection{Planner Scalability and Real-time Responsiveness}
\label{sec:stress-test}
Next, we solely evaluate the planner's capability while changing the number of \emph{simulated} robots to test its scalability.
We use identical parameters as the ones in our physical deployments:
for each team size, up to 64, we prepared 100 random instances within \qtyproduct[product-units=bracket-power]{6 x 6 x 2}{\metre} similar in size to our arena, with five pole-shaped obstacles of radius \SI{0.2}{\meter}.
The robot radius was set to \SI{0.2}{\meter}.

\Cref{fig:planner_test} summarizes the results in terms of the planner's ability to find a solution from scratch within a given time limit.
With moderate number of robots ($\sim$$32$), the planner reliably obtains solutions under \SI{200}{\milli\second}, enabling us to run MAPF at least $\SI{5}{\hertz}$ stably.
This consequently allows us to run MAPF at \SIrange{10}{20}{Hz} when operating swarms with practical sizes in our arena ($\leq$$16$).
As expected, as the number of robots increases, it becomes more difficult to find solutions in the short time available.
However, we note that the difficulty for the planner comes from the density of the configuration rather than the number of robots.
For instance, when the robot radius is set to \SI{0.1}{\meter} for the same problem, the planner stably obtains solutions for 100 robots in \SI{250}{\milli\second}.
An extreme case is shown in the supplementary video, where the planner derives a solution for 500 robots in about \SI{3.5}{\second};
this scalability underpins \LF's rapid replanning for reasonably-sized robot teams.

As illustrated in \cref{fig:planner_anytime}, the \textit{quality} of the solutions improves over time.
Quantitative results are available in \cref{fig:planner_test}.
In practice, \LF{} reuses previous solutions to further boost the refinement process.

\subsection{Juxtaposing with Decoupled Method}
Next, we showcase the performance of \LF{} in terms of its reliability in finding solutions, and the quality of the trajectories executed by it -- and contrast it against the solutions found by AMSwarm.
As outlined in \cref{sec:target_sys}, \LF{} is a \textit{coupled} approach, and is targeted towards scenarios where full information about the team is available.
In contrast, decoupled approaches such as AMSwarm attempt to solve individual optimization problems `locally' using the team information available at a given time.
This is generally faster since agents only need to solve a smaller problem, often over a receding horizon.
We use the default configurations in AMSwarm, with some minor adjustments to tune the parameters comparable to \LF's defaults, such as arena size, smoothness order, and velocity limits.

\Cref{fig:juxt} depicts 12 manually-created test-cases (6 `easy' and 6 `hard') involving five obstacles and five agents, with trajectories executed by AMSwarm and \LF{} overlaid.
We consider a configuration `easy' if it involves very few potential conflicts, while `hard' configurations are ones that induce several simultaneous conflicts.
We observe that \LF{} is able to successfully navigate all agents towards their destinations in all configurations, while AMSwarm fails to in certain hard problems.
The failure cases are most often ``livelocks'', where several agents get stuck in a local optimum and successively pick actions that cause conflicts for another.
In the easy configurations, the solutions executed by \LF{} have a comparable total path length, albeit slightly higher than AMSwarm's executed trajectories.
The computation time (not shown) with \LF{} is an order of magnitude higher,
however, the mission time is generally lower since \LF{} produces trajectories with higher accelerations.

\subsection{Additional Case Studies}
The following is intended to demonstrate \LF's flexibility in various scenarios involving physical robots and dynamic obstacles.
The supplementary videos include these, and additional deployments with up to eight ground robots~\cite{blumenkamp2024cambridge}.

\subsubsection{Asynchronous Missions}
Since \LF replans coordinated paths at high frequency, it naturally adapts to an \emph{asynchronous} scenario where an agent is assigned a new goal location immediately after reaching the current one.
This is in contrast to the \emph{synchronous} scenario presented in \cref{sec:navigation_performance}, where a new target assignment for the entire team is made only after all robots have reached their current targets.
In the MAPF literature, the asynchronous style is also called a lifelong scenario~\cite{li2021lifelong}.
The adaptation is simple; \LF replans the paths for all the robots when one of the goals is updated.

\Cref{fig:sync_async}, using the same mission generation scheme with the ten-robot case in \cref{sec:navigation_performance}, highlights the differences between synchronous and asynchronous mission execution with \LF.
The asynchronous scenario results in slightly larger tracking errors because it forces the planner to completely replan more often than in the synchronous case;
i.e. it takes time for the paths to become kinodynamically smooth.
Meanwhile, as shown in the corresponding scatter plots (\cref{fig:sync_async}, right), with an enforced synchronous style, agents with shorter goal distances need to wait the same amount of time as the rest of the team.
In contrast, asynchronous replanning enables a more proportional distribution.

\input{figs/sync_async}
\input{figs/following}
\input{figs/15drones}

\subsubsection{Target Following with Ground Robots}
\Cref{fig:follower} presents a different type of \LF demonstration, where four ground robots follow a manually controlled box-wearing robot, maintaining a circular shape around the target, while avoiding static obstacles and inter-agent collisions.
\LF seamlessly handles this task by specifying a set of target locations around the target robot for each replanning.
This shows \LF{}'s adaptability to different tasks and robot platforms.

\subsubsection{Multirotor Swarm Deployment with Pedestrian}
Finally, we deploy 15 multirotors in the synchronous random target scenario while a human, whose position is tracked by the motion capture system, walks through the swarm.
\LF safely controls the swarm in its continued random-goal mission, avoiding collisions between them and with a human.
This demonstration showcases \LF's ability to handle a large team, even in dynamic environments.

\section{Conclusion}
We presented \LF, a framework based on the belief that combining fast MAPF with robot-wise optimal trajectory control provides a powerful multi-robot control scheme.
Our motivation is to integrate the high-level planner closely with the operation of an optimal on-robot controller in an MPC-like fashion.
The ingredients for doing so are LaCAM for the MAPF and Freyja for the controller, but the concept itself lends to various implementations.
For example, we consider that LaCAM can be replaced by other sufficiently fast methods~\cite{li2022mapf}, which is an active area of research.
With their future developments, incorporating a coupled high-level planner into the feedback loop will be more robust and computationally efficient to derive near-optimal coordination.

\textit{Limitations and Further Work:}
Although the framework's concept is an attractive option for multi-robot control, we acknowledge two key technical limitations.
These include the inability to perform agile maneuvers that are harder to approximate using a single integrator model used in MAPF.
Likewise, we currently do not accommodate any acceleration shaping in the planner or the controller.
Overcoming this requires a trade-off with scalability and real-time planning ability, as the planner's state space increases with an extended state space representation for velocity and acceleration profiles.
Another limitation is the need for complete environmental knowledge to carry out coupled planning.
We are interested in relaxing this assumption, i.e., a control scheme that works with partially observable environments, potentially with the assistance of data-driven methods~\cite{li2021message}.

%% file: figs/top.tex
{
\begin{figure}[t!]
\centering
\includegraphics[width=0.98\linewidth]{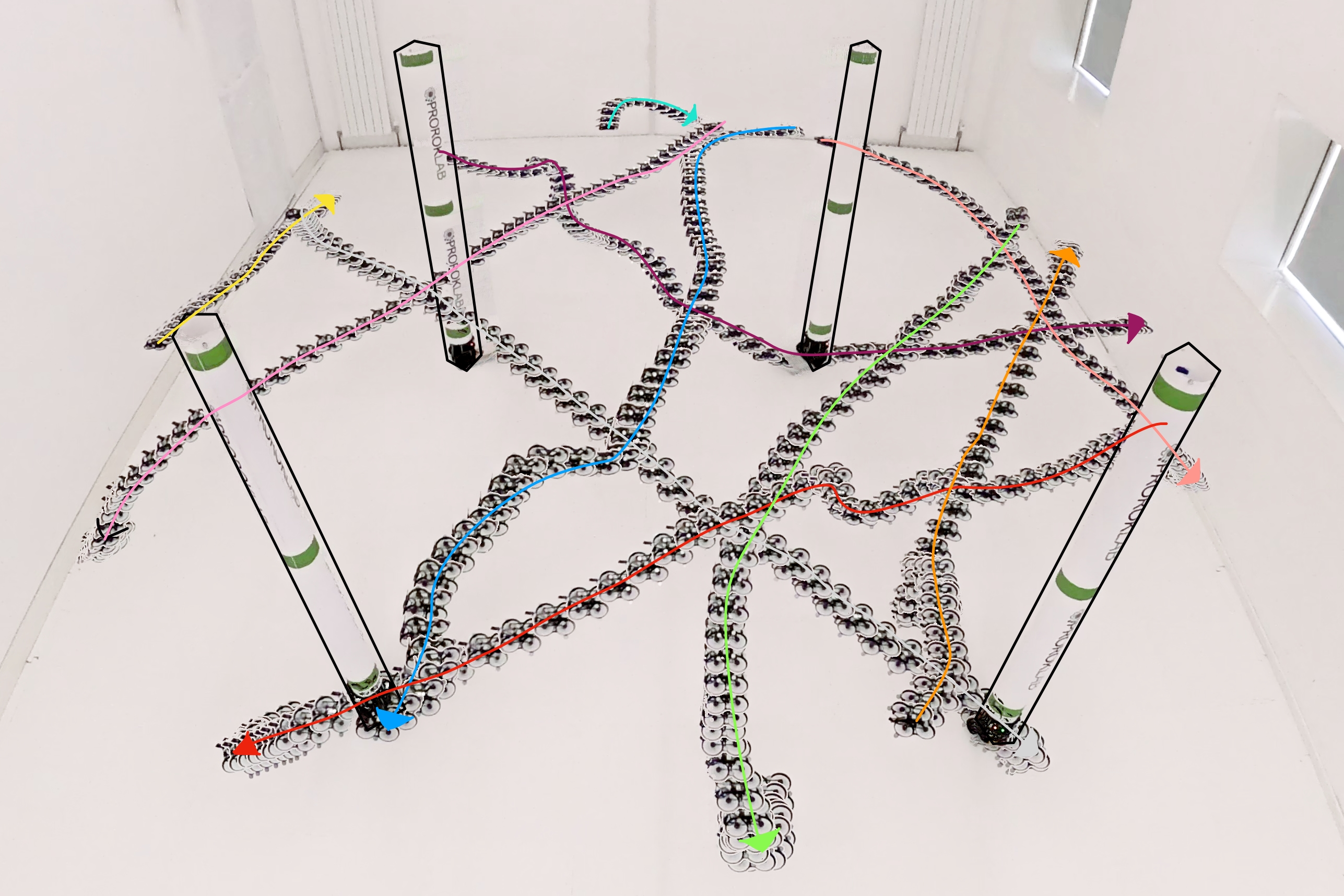}
\caption{
Navigation trajectories of ten multirotors with four dynamic obstacles by \lf.
Obstacles are highlighted with black outlines.
}
\label{fig:top}
\end{figure}
}

%% file: table/characterization.tex
{
\newcommand{\block}[1]{{\setlength{\tabcolsep}{0pt}\begin{tabular}{l}#1\end{tabular}}}
\renewcommand{\c}{\m{\checkmark}}
\newcommand{\rc}{\rowcolor[gray]{.9}}
\renewcommand{\r}[1]{\rotatebox{90}{\hspace{-0.2cm}{#1}}}
\setlength{\tabcolsep}{4pt}
\begin{table}[t!]
\centering
\caption{
Characterization on centralized planning styles.
}
\label{table:characterization}
\begin{tabular}{lllllllll}
\toprule
\block{planning\\horizon} & \block{when to\\plan} & \block{represen-\\tation} & examples &
\r{scalability} & \r{adaptivity} & \r{tight coord.} & \r{long coord.} & \r{smoothness}
\\\midrule
full & oneshot & coupled &
\cite{augugliaro2012generation,honig2018trajectory} &
& & \c & \c & \c
\\\rc
limited & continual & coupled &
\cite{soria2021predictive,tajbakhsh2024conflict} &
& \c & \c &  & \c
\\
limited & continual & decoupled &
\cite{adajania2023amswarm,adajania2024amswarmx} &
\c & \c & & & \c
\\\rc
\textbf{full} & \textbf{continual} & \textbf{coupled} &
\textbf{ours (\LF)} &
\c & \c & \c & \c &
\\\bottomrule
\end{tabular}
\end{table}
}

%% file: figs/arch.tex
\begin{figure}
    \centering
    \includegraphics[width=\linewidth]{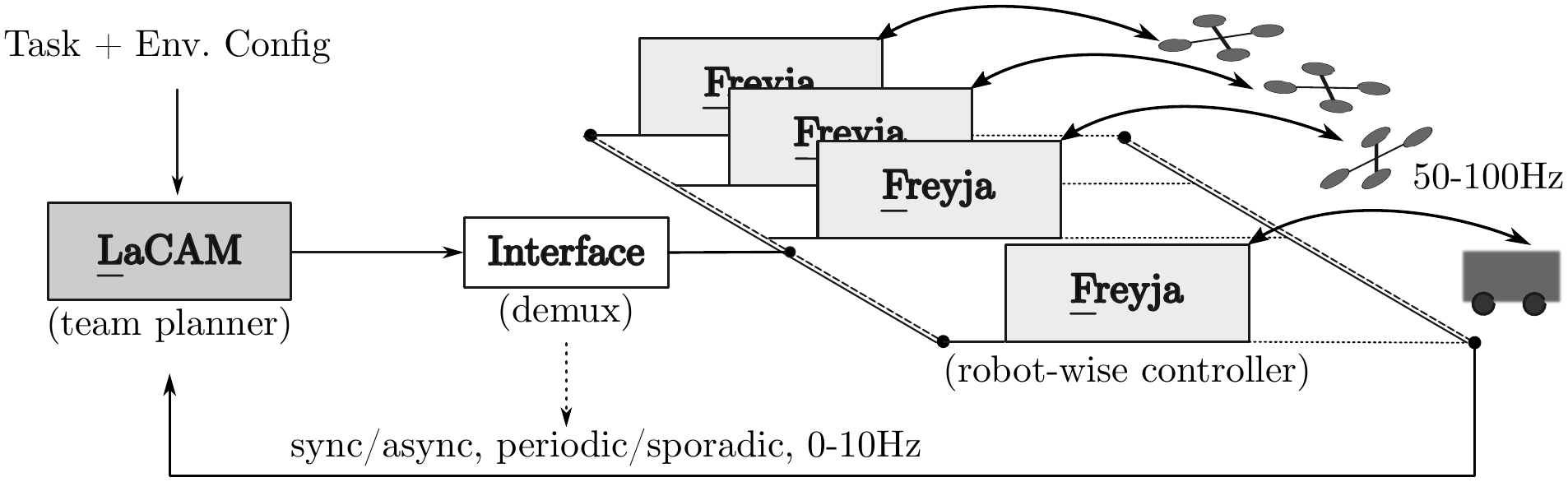}
    \caption{A schematic view of \LF's architecture, blending LaCAM as a fast team-level planner and Freyja as a robust on-robot controller. The demultiplexing interface simply reformats planner outputs, and is also capable of invoking it if required.}
    \label{fig:arch}
\end{figure}

%% file: figs/navigation_performance.tex
\begin{figure*}[tp!]
    \centering
    \begin{subfigure}[b]{0.32\textwidth}
        \centering
        \includegraphics[width=\linewidth]
            {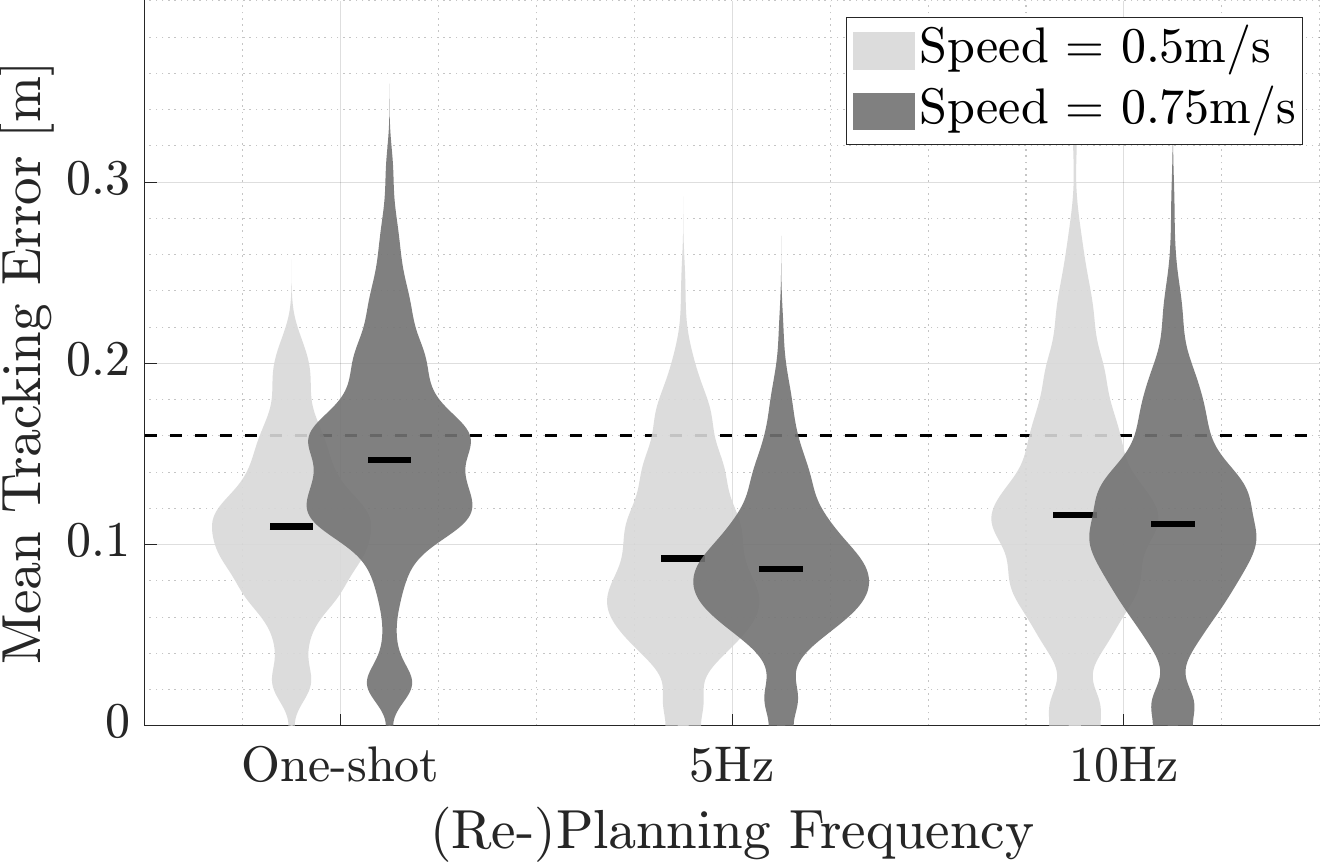}
        \caption{5 robots, no obst.; replanning performance}
        \label{fig:replanning-freq-comp}
    \end{subfigure}
    \begin{subfigure}[b]{0.32\textwidth}
        \centering
        \includegraphics[width=\linewidth]
            {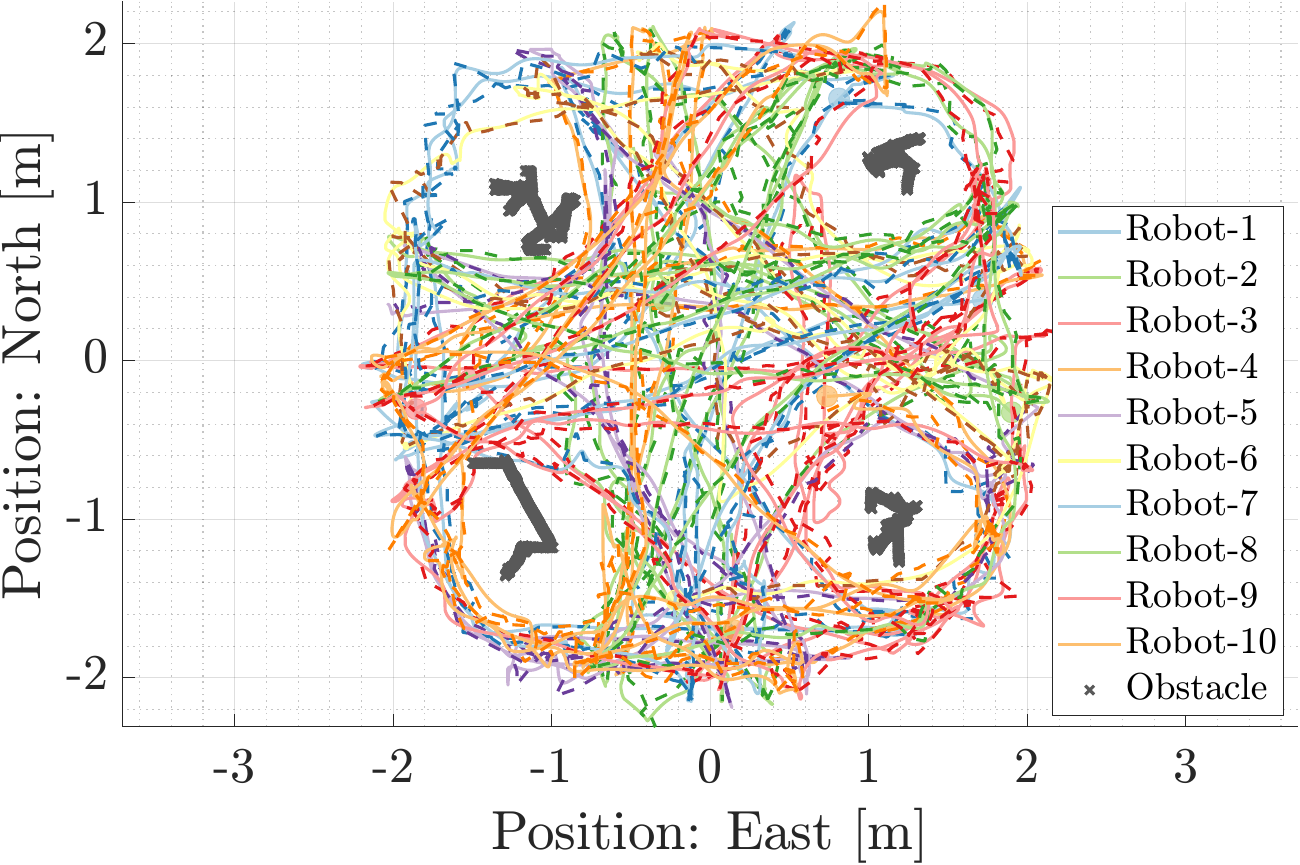}
        \caption{10 robots, 4 dynamic obstacles}
        \label{fig:10rob-obst-topdown}
    \end{subfigure}
    \begin{subfigure}[b]{0.32\textwidth}
        \centering
        \includegraphics[width=\linewidth]
            {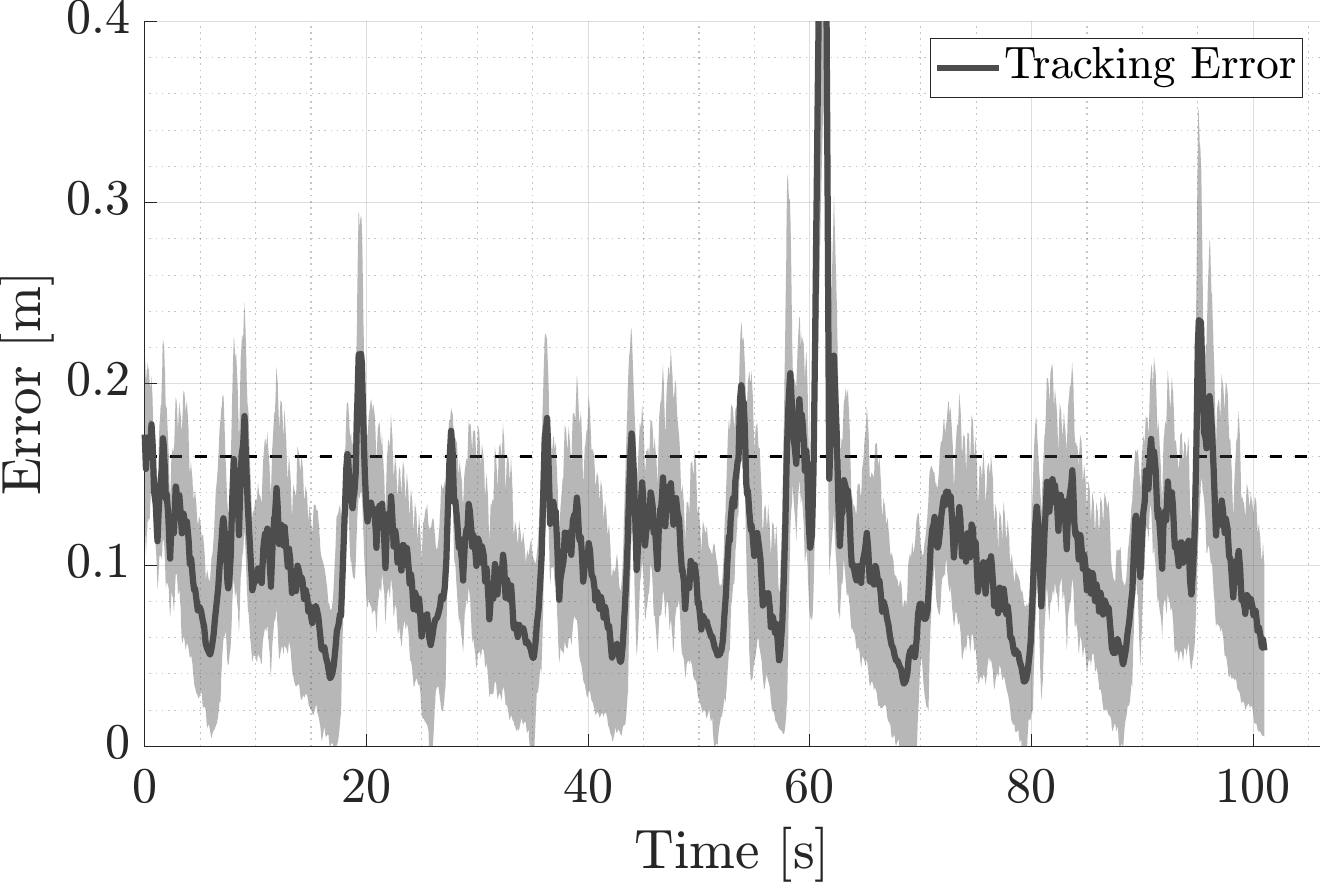}
        \caption{10 robots, 4 dynamic obstacles}
        \label{fig:10rob-obst-perf}
    \end{subfigure}
    \caption{Real-world navigation performance with 5 and 10 robots, with/without obstacles, and different replanning frequencies. The dashed black-line indicates one robot diameter.}
    \label{fig:navigation_performance}
\end{figure*}

%% file: figs/planner_anytime.tex
{
\begin{figure}[t!]
\centering
\includegraphics[width=0.24\linewidth]{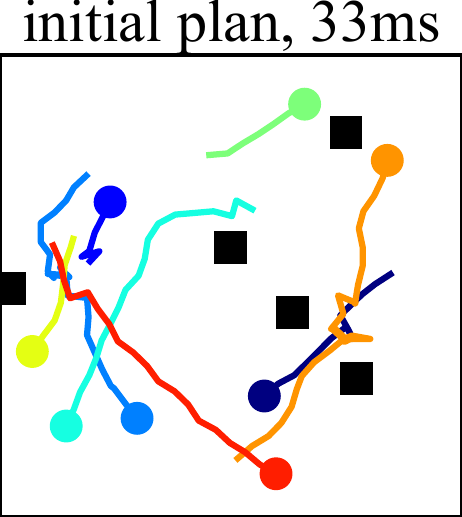}
\includegraphics[width=0.24\linewidth]{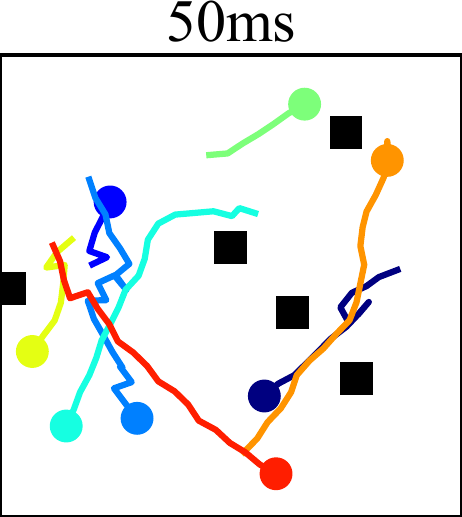}
\includegraphics[width=0.24\linewidth]{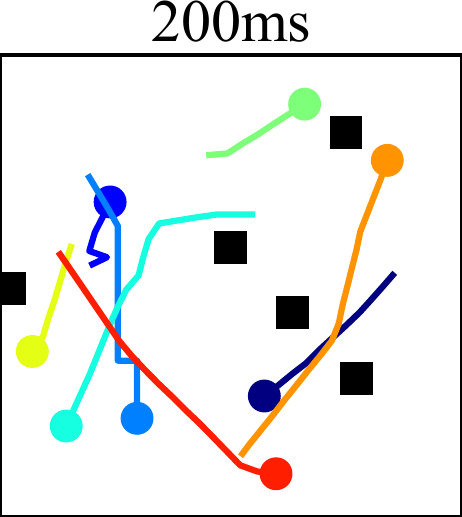}
\includegraphics[width=0.24\linewidth]{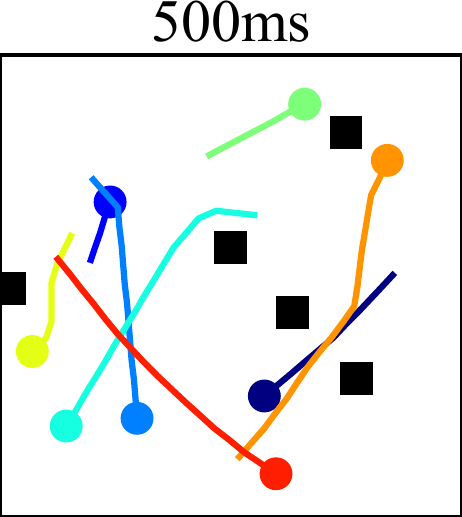}
\caption{
  Anytime planning visualization.
  The trajectories are 2D projections (XY coordinates) of one instance in \cref{fig:planner_test}.
}
\label{fig:planner_anytime}
\end{figure}
}

%% file: figs/planner_test.tex
{
\begin{figure}[tp!]
\centering
\includegraphics[width=0.49\linewidth]{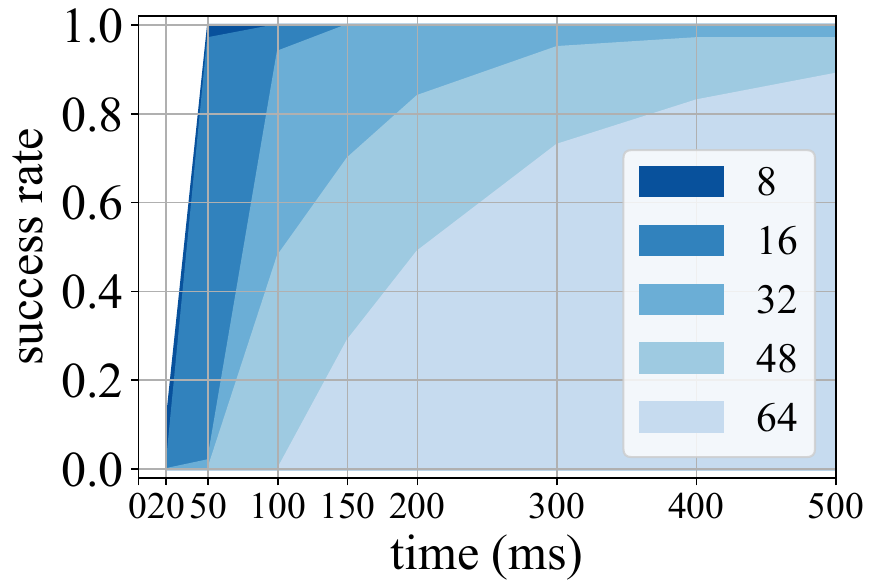}
\includegraphics[width=0.49\linewidth]{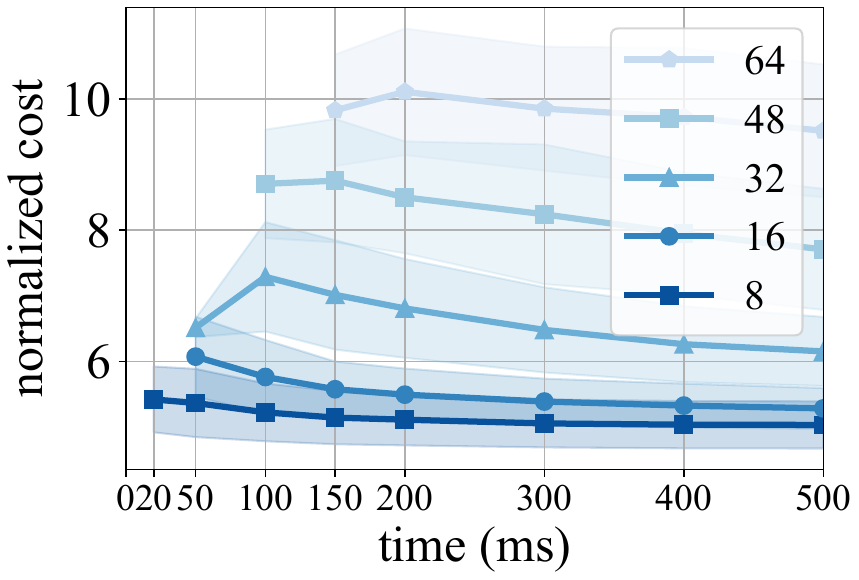}
\caption{
  Real-time planning capability with different number of agents.
  The `cost' on the right displays the average flowtime with standard deviation, i.e., the sum of the travel time normalized by the start-goal distance.
}
\label{fig:planner_test}
\end{figure}
}

%% file: figs/juxt.tex
\begin{figure*}
    \centering
    \includegraphics[width=\linewidth]{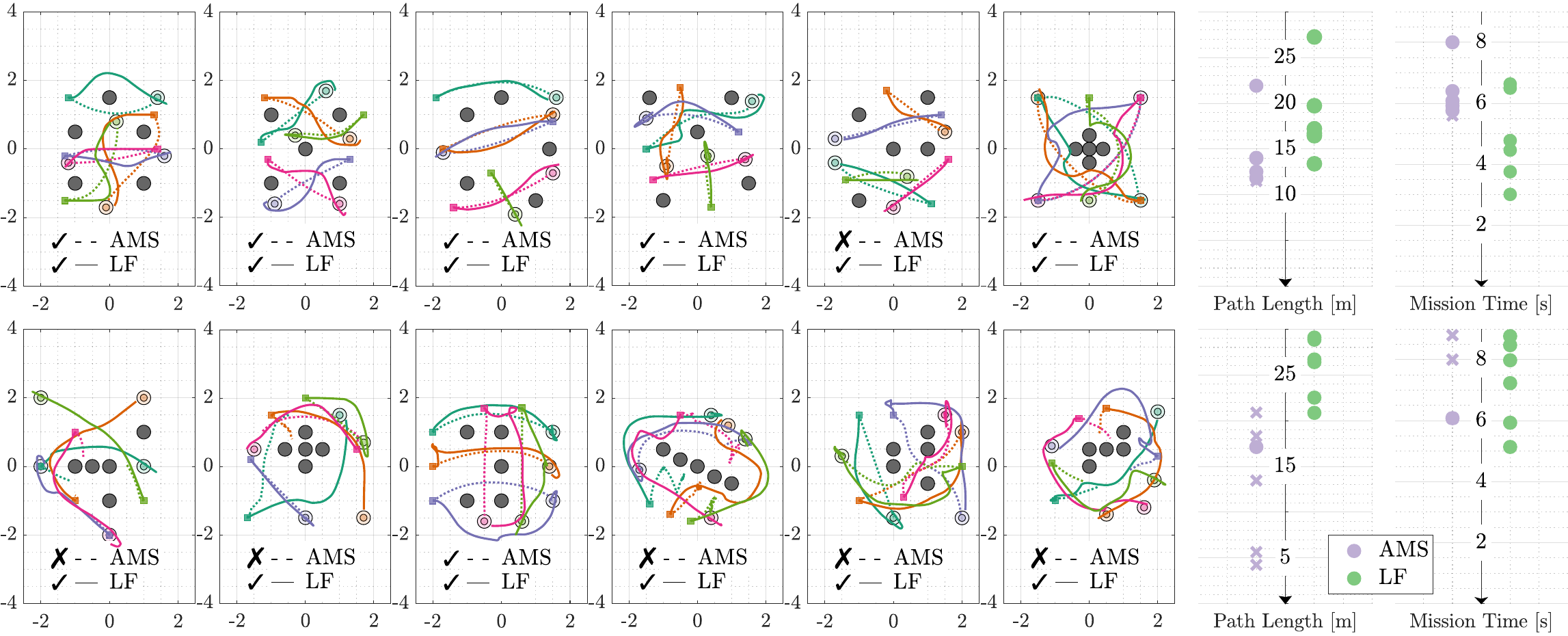}
    \caption{Juxtaposing \LF{} with a numerical optimization-based decoupled approach, AMSwarm (``AMS")~\cite{adajania2023amswarm} in various simulated problems involving five obstacles (\textcolor{black!70}{$\bullet$}) and five agents ($\textcolor[rgb]{0.16,0.62,0.46}{\wr}\,
    \textcolor[rgb]{0.85,0.37,0.01}{\wr}\,
    \textcolor[rgb]{0.46,0.44,0.70}{\wr}\,
    \textcolor[rgb]{0.91,0.16,0.54}{\wr}\,
    \textcolor[rgb]{0.40,0.65,0.12}{\wr}$).
    The top row contains `easy' scenarios that force 1-2 robots to deconflict their trajectories, while the bottom row contains `harder' scenarios that force several robots to solve a highly coupled problem. \LF{} generally produces paths with comparable lengths, and with a lower total mission time. In the hard scenarios, \LF{} can find successful solutions in all cases (\checkmark), whereas AMSwarm often gets stuck in a local solution that is infeasible when multiple agents are solving their decoupled problems ($\times$).
    }
    \label{fig:juxt}
\end{figure*}

%% file: figs/sync_async.tex
{
\begin{figure}
\centering
\scalebox{1.0}{
\begin{tikzpicture}
\node[anchor=south west] at (0, 0)
  {\includegraphics[width=0.98\linewidth]{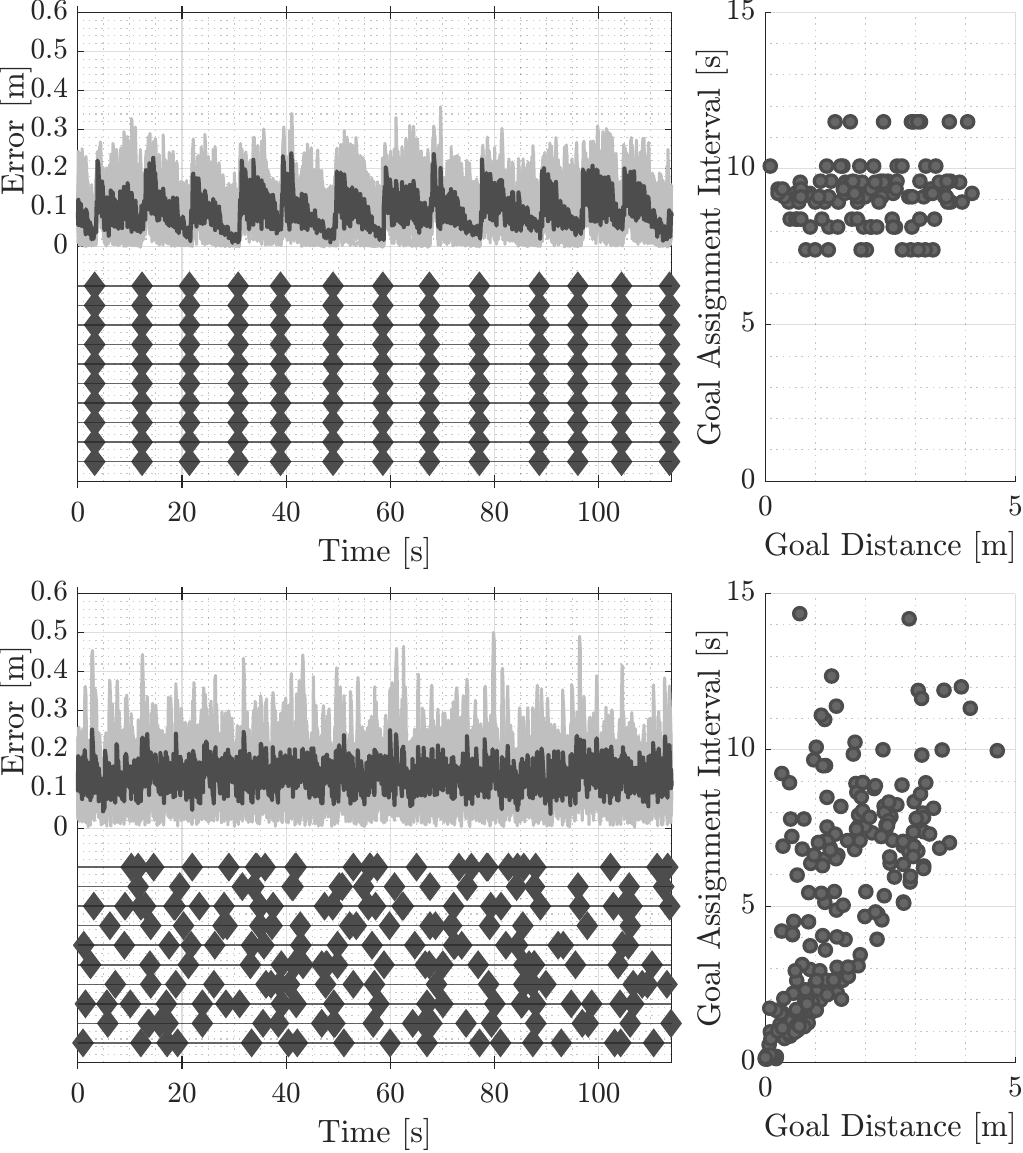}};
\node[anchor=south west, rotate=90] at (0.7, 6.1) {\scriptsize Sync};
\node[anchor=south west, rotate=90] at (0.7, 1.3) {\scriptsize Async};
\end{tikzpicture}
}
\caption{
Process visualization of synchronous (upper) and asynchronous (lower) missions, with ten real drones and four dynamic obstacles deployed as \cref{fig:top}.
The left figures show the tracking error and when the new target was assigned to each robot.
The figures on the right analyze each task in terms of the distance from start to goal and the time duration until replanning for a new goal.
}
\label{fig:sync_async}
\end{figure}
}

%% file: figs/following.tex
{
\begin{figure}[t!]
\centering
\scalebox{1.0}{
\begin{tikzpicture}
\node[anchor=south west] at (0, 0) {\includegraphics[width=1.0\linewidth]{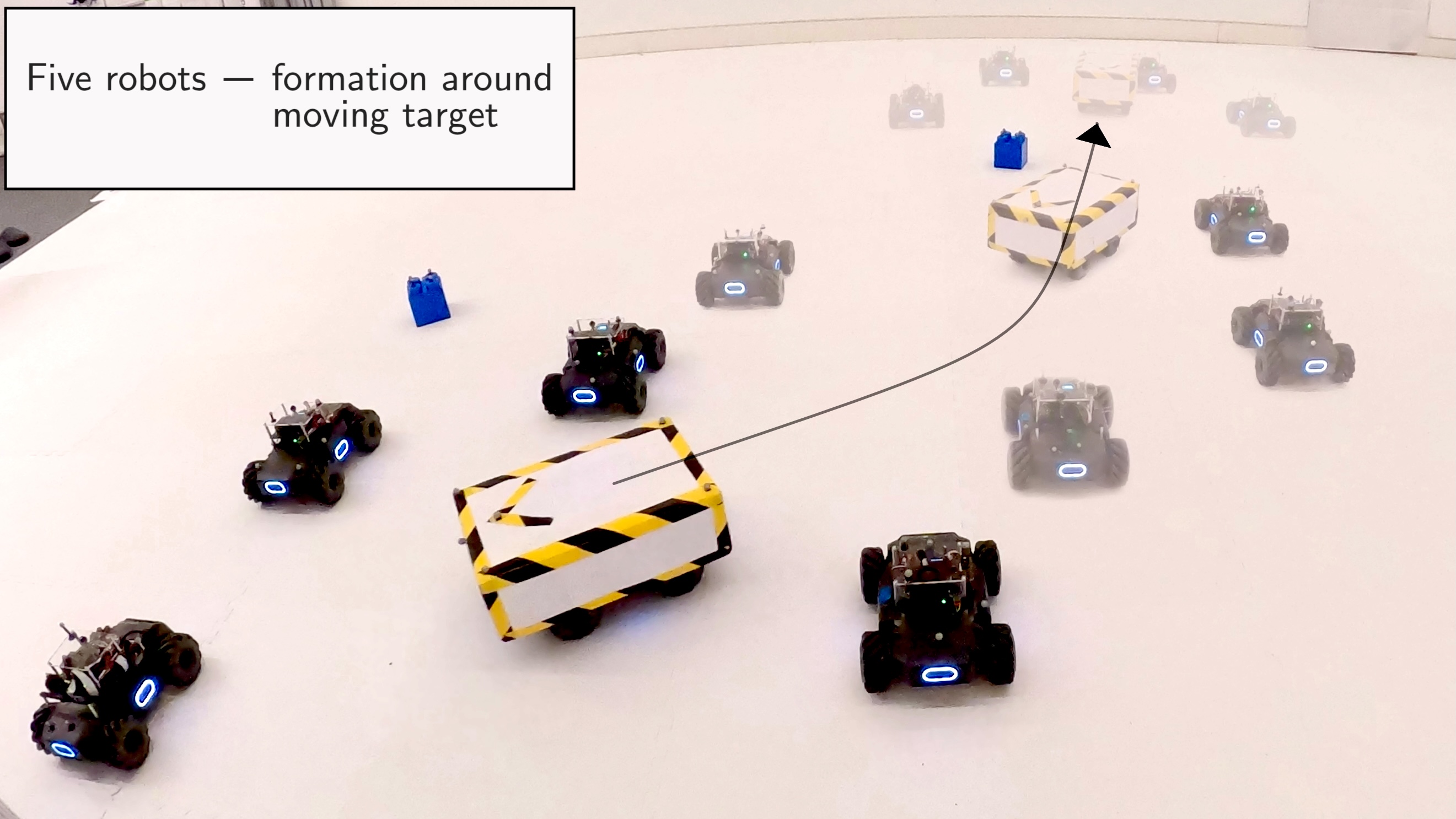}};
\node[anchor=south west] at (2.8, 0.6) {target};
\end{tikzpicture}
}
\caption{
Target following demonstration where four ground robots follow the box-wearing robot which is manually controlled. The image is made up of three superimposed snapshots.
}
\label{fig:follower}
\end{figure}
}

%% file: figs/15drones.tex
{
\begin{figure*}[t!]
\centering
\newcommand{\imgwidth}{0.32\linewidth}
\scalebox{1.0}{
\begin{tikzpicture}
\node[anchor=south west] at (0, 0) {
\includegraphics[width=\imgwidth]{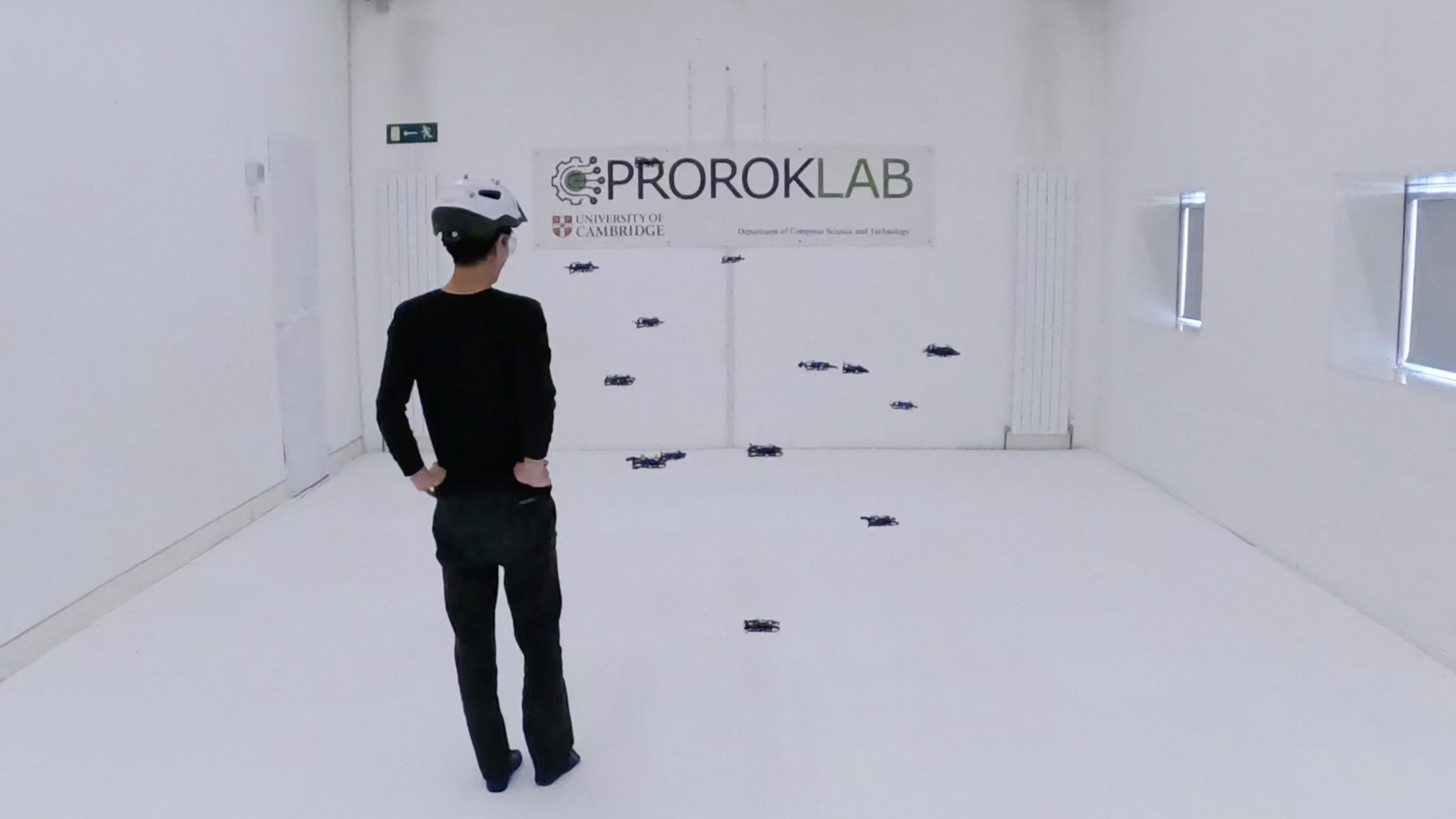}
\includegraphics[width=\imgwidth]{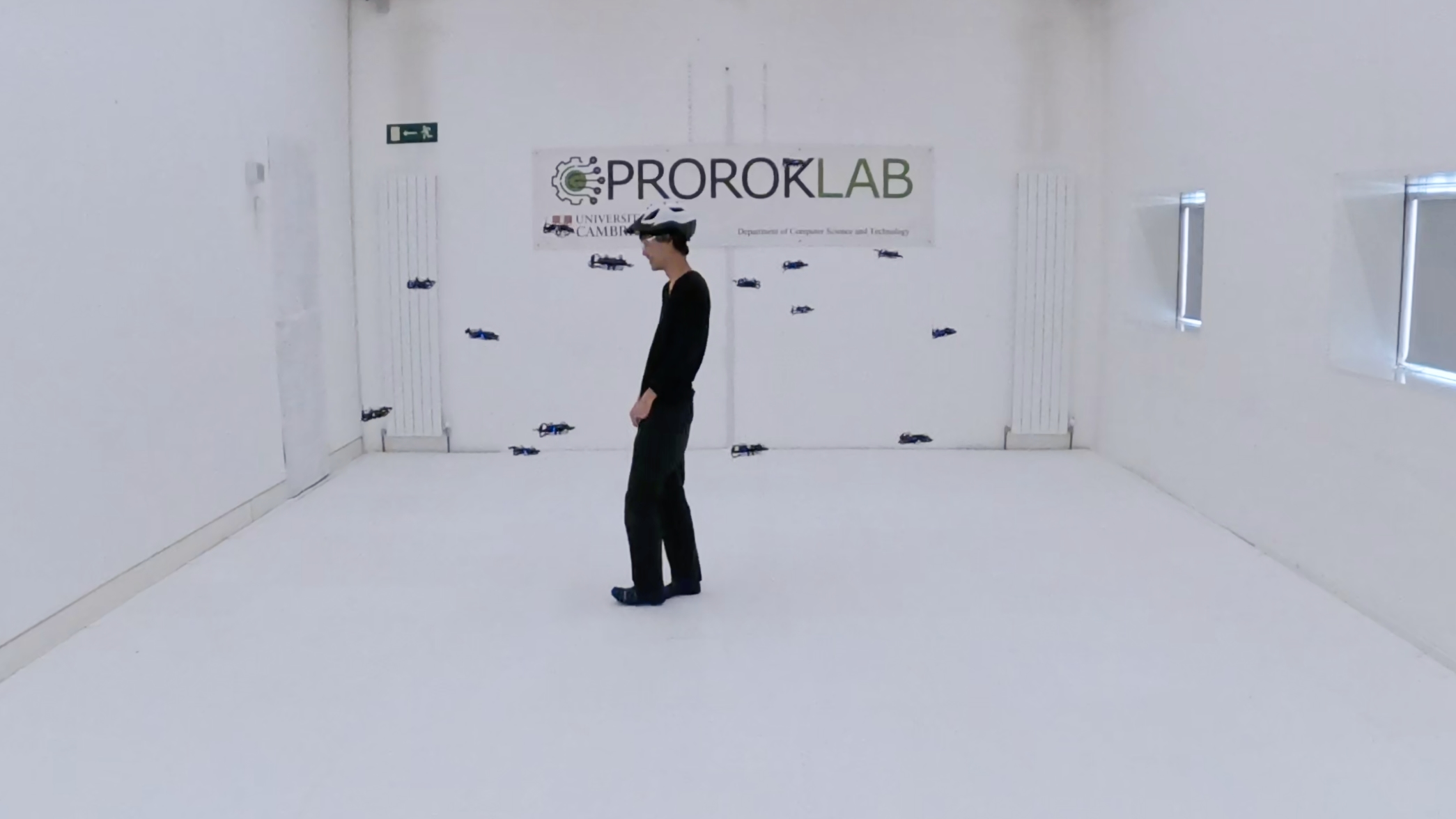}
\includegraphics[width=\imgwidth]{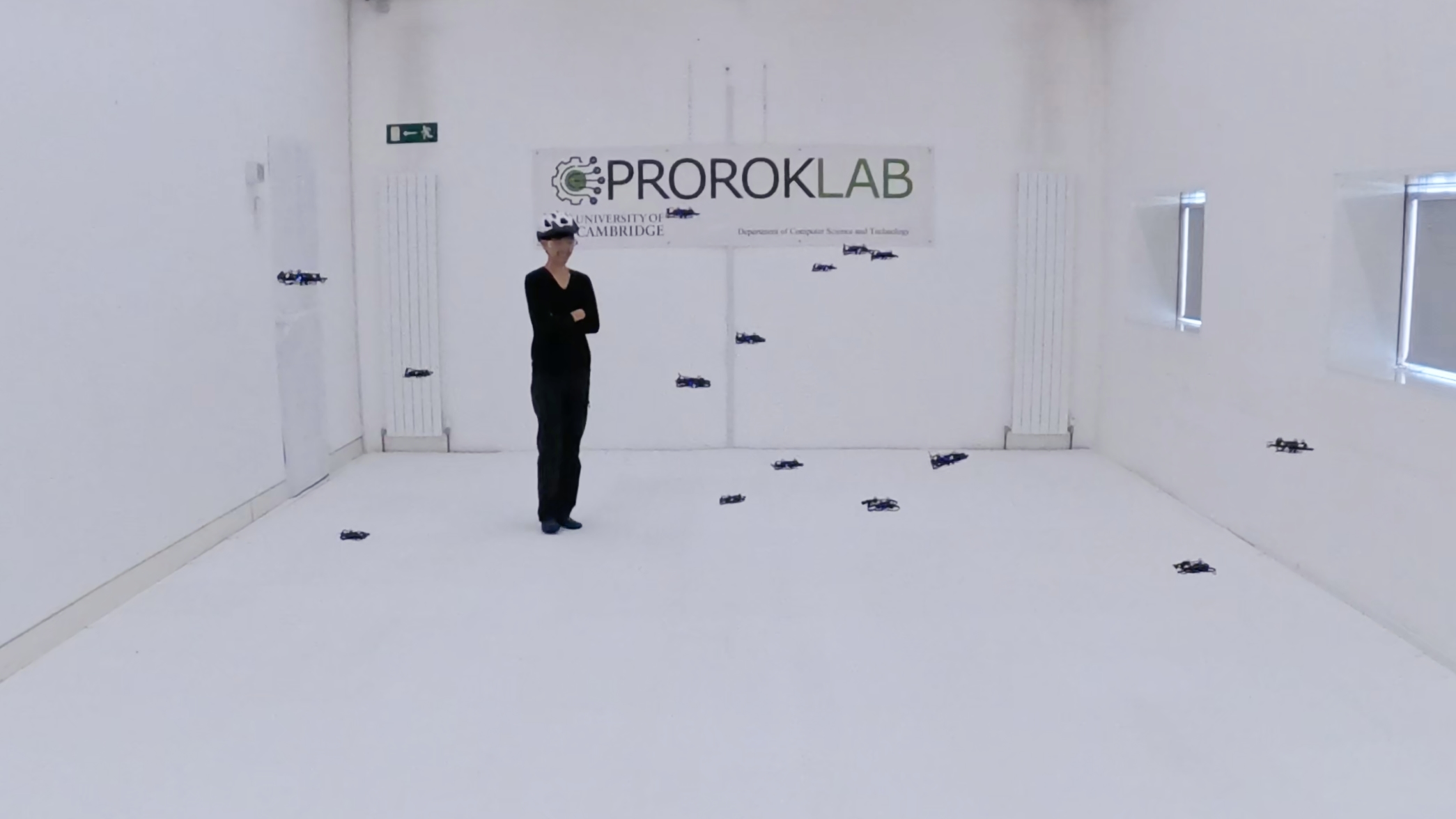}
};
\draw[very thick,->] (5.3,0.5) --+ (1.0, 0);
\draw[very thick,->] (11.2,0.5) --+ (1.0, 0);
\end{tikzpicture}
}
\caption{
Snapshots of a person walking from front to back ends, among 15 drones that are continuously solving synchronous missions.
}
\label{fig:15drones}
\end{figure*}
}